\documentclass{WileyMSP-template}

\usepackage{xcolor}
\usepackage{colortbl}
\usepackage{booktabs}
\usepackage{multirow}
\usepackage{array}
\usepackage{float}
\usepackage[hidelinks]{hyperref}
\usepackage{graphicx}

\begin{document}

\pagestyle{fancy}
\rhead{\includegraphics[width=2.5cm]{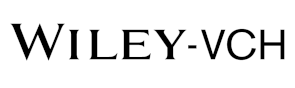}}

\title{Integrating Large Language Models and Knowledge Graphs to Capture Political Viewpoints in News Media}

\maketitle

% Author: Please give full first and last names for authors and include * after the name of all corresponding authors

\author{Massimiliano Fadda}
\author{Enrico Motta}
\author{Francesco Osborne} 
\author{Diego Reforgiato Recupero}
\author{Angelo Salatino}

% Dedication

%\dedication{Optional dedication here. If no dedication is required, please leave blank}

% Affiliations: Please provide adacemic titles (Prof. or Dr.) for all authors where applicable, and include an institutional email address for all corresponding authors
\begin{affiliations}
M. Fadda, D. Reforgiato Recupero\\
Department of Mathematics and Computer Science, University of Cagliari, Via Ospedale 72, 09124, Cagliari, Italy\\
E. Motta, F. Osborne, A. Salatino\\
Knowledge Media Institute, The Open University, Walton Hall, Milton Keynes, MK7 6AA UK\\
\end{affiliations}

% Keywords: Please provide a minimum of three and a maximum of seven keywords, separated by commas


% Abstract should be written in the present tense and impersonal style (i.e., avoid we), and be at most 200 words long
\begin{abstract}

News sources play a central role in democratic societies by shaping political and social discourse through specific \textit{topics}, \textit{viewpoints} and \textit{voices}. Understanding these dynamics is essential for assessing whether the media landscape offers a balanced and fair account of public debate. In earlier work, we introduced a pipeline that, given a news corpus, i) uses a hybrid human–machine approach to identify the range of viewpoints expressed about a given topic, and ii) classifies relevant \textit{claims} with respect to the identified viewpoints, defined as sets of semantically and ideologically congruent claims (e.g., positions arguing that immigration positively impacts the UK economy).
In this paper, we improve this pipeline by i) fine-tuning Large Language Models (LLMs) for viewpoint classification and ii) enriching claim representations with semantic descriptions of relevant actors drawn from Wikidata. We evaluate our approach against alternative solutions on a benchmark centred on the UK immigration debate. Results show that while both mechanisms independently improve classification performance, their integration yields the best results, particularly when using LLMs capable of processing long inputs.
%News sources play a central role in democratic societies: they shape political and social discourse by giving a platform to specific \textit{topics}, \textit{viewpoints} and \textit{voices}. Understanding the dynamics of these representations is essential for assessing whether the media landscape offers a balanced and fair account of public debate. In our earlier work we introduced an effective pipeline that, given a news corpus, i) first uses a hybrid human-machine approach to identify the range of viewpoints expressed in the corpus about a given topic, and ii) is then able to classify effectively relevant \textit{claims} expressed about the topic in question, with respect to the identified viewpoints. The latter are defined as sets of claims that are semantically and ideologically congruent — e.g., all positions that argue that  immigration has a positive impact on the UK's economy. 
%In this paper we improve on our earlier pipeline by i) fine-tuning Large Language Models (LLMs) for the viewpoint classification task and ii) enriching the representation of claims by means of a semantic description of the relevant actors, which is drawn from Wikidata. We evaluate our approach against alternative solutions using a benchmark centred on the immigration debate in the UK. The evaluation indicates that while both of these mechanisms are individually able to improve performance in the claim classification task, their integration provides the best results, particularly when leveraging an LLM capable of processing substantial input sizes.
\end{abstract}

% Text: Please use section headings and subheadings as specified below. For communications, all section headings apart from Experimental Section should be removed
% Please make the first reference to a display item bold: \textbf{Figure 1}
% Do not abbreviate Figure, Equation, etc.; display items are always singular, i.e., Figure 1 and 2.
% Equations are always singular, i.e., Equation 1 and 2, and should be inserted using the {equation} environment, not as graphics
% Please do not use footnotes in the text, additional information can be added to the Reference list.

\section{Introduction}\label{sec1}
News sources play a crucial role in democratic societies by highlighting specific \textit{topics}, \textit{viewpoints} and \textit{voices} in their representation of political and social discourse. Understanding the dynamics of these representations is essential for evaluating to what extent the media landscape provides a balanced and fair account of the debate around a topic, thus fulfilling its role to allow citizens to make informed decisions on the basis of their access to a variety of viewpoints~\cite{Smith01072012,trappel2011media}. 
In our earlier work~\cite{motta2024}, we introduced a hybrid human-machine approach to assessing  \textit{viewpoint diversity}~\cite{Masini18112018}.  In particular, given a news corpus, our pipeline i) first identifies the range of viewpoints expressed in the corpus about a given topic, and ii) then classifies relevant \textit{claims} expressed about the topic in question, with respect to the identified viewpoints. A claim is formally defined as a sextuple \textlangle utterance, actor, news item, news source,
date, topic\textrangle, where the utterance is a statement made by an actor, about a particular
topic, which is reported in a news item, which was published by a news source on a certain date~\cite{motta2024}. In turn, viewpoints are defined as sets of claims that are semantically and ideologically congruent — e.g., all positions that argue that  immigration has a positive impact on the UK's economy.

In this paper we extend our pipeline by introducing two key improvements. Specifically, we fine-tune Large Language Models (LLMs) for the claim classification task and we enrich the claim representation with a semantic description of the relevant actors, which is drawn from Wikidata~\cite{vrandevcic2023wikidata} and represented through a Knowledge Graph (KG)~\cite{peng2023knowledge}. We then evaluate our approach against alternative solutions, using a benchmark centred on the immigration debate in the UK. In particular, we analyse different approaches to contextualising a claim (using the surrounding text, leveraging structured information from the KG and combining both) and evaluate their impact on the performance of the claim classification engine. The evaluation indicates that, while both fine-tuning and the adoption of a KG are able to improve performance individually, their integration provides the best results (91.7\% F1 score), particularly when leveraging an LLM capable of processing substantial input sizes.

To ensure full reproducibility, all materials have been released, including both the evaluation dataset and the LLM prompts.\footnote{Anonymous GitHub repository — \url{https://anonymous.4open.science/r/Political_Viewpoints_in_News_Media-3BE6/README.md}}

The rest of the paper is organised as follows. In Section~\ref{sec:related_work}, we discuss the relevant literature, focusing in particular on computational solutions for news analytics and classification approaches that combine LLMs and KGs. In Section~\ref{sec:methodology}, we present our methodology. %, which extends the approach developed by~\cite{motta2024} by introducing a KG which covers relevant background information about the actors expressing claims reported in the news.
In Section~\ref{sec:eval}, we report the results from our experiments. %, in particular showing that the combination of an LLM and the KG improves the performance of the claim classification module. 
Finally, in Section~\ref{sec:conclusions}, we comment on the results presented in this work and discuss future directions of research. %the issues that will be the focus of future research.

\section{Related work}\label{sec:related_work}

In this section, we review the literature relevant to this paper, with a particular focus on the primary task of interest (classifying claims based on viewpoints) and the architectural approach we have adopted (integrating LLMs with KGs).

%In this section we report on the literature that is relevant to the discussion in this paper, focusing in particular on the main task we are concerned with (classifying claims with respect to viewpoints) and the architectural solution we have adopted (integration of LLMs and KGs).

\subsection{Automatic viewpoint capture}

As already mentioned, the work presented here builds on an earlier paper of ours~\cite{motta2024}, which introduces a semi-automated approach to detecting viewpoints relevant to a topic and then uses a LLM in a zero-shot setting to classify the claims extracted from a news corpus with respect to the relevant viewpoints. While the obtained results are satisfactory, in this follow-up paper we show that a performance improvement can be achieved by  introducing LLM fine-tuning and enriching the claim representation through an automatically-generated semantic description of the associated actors.

Chen et al.~\cite{chen2022toward} address the automatic discovery of diverse perspectives and propose a framework that has a number of commonalities and differences with the approach taken by us.  In particular, they focus on retrieving perspectives that are relevant to a claim and consider a number of tasks: identifying the stance of these perspectives with respect to the original claim; identifying equivalent perspectives; and, given a claim and a perspective, identifying evidence for this stance from the associated corpus.  They also show that the RoBERTa Large model can perform reasonably well on these tasks. This approach has a number of similarities with ours: their identification of equivalent perspectives is analogous to our definition of a viewpoint as a collection of congruent claims. They also identify perspectives on the basis of a given claim, which is similar to our approach that identifies the claims relevant to a given topic. Having said so, we would argue that the conceptual framework underlying our approach provides a more robust basis for the realization of news analytics solutions than the one adopted by Chen et al. In particular, in their approach, there is no explicit characterization of a topic and the process of identifying relevant perspectives is driven by the selection of a particular claim. Here, it could be argued that it is unclear what is the epistemological basis for distinguishing between a claim and a perspective.  More importantly, while a claim implies a topic, the lack of an explicit characterization of the current topic makes it difficult to map the space of viewpoints  to the topic in question (which is implied by the selected claim).  Indeed, in contrast with our approach, there is no attempt in~\cite{chen2022toward} at providing such a characterization of the space of viewpoints. On the contrary, topics, claims and viewpoints play clearly distinctive roles in our framework, which is formally analysed and represented as an ontology in~\cite{motta2025epistemology}. Another difference between this work and that reported by Chen et al. concerns the architecture of the classifier used for the task of identifying equivalent perspectives – i.e., classifying a claim with respect to the available set of viewpoints in our framework.  In our approach we combine LLM fine-tuning and a semantic representation of the actors, in contrast with Chen et al., who only consider model fine-tuning. 
Chen et al.~\cite{chen2022toward} address the automatic discovery of diverse perspectives and propose a framework that has a number of commonalities and differences with the approach taken by us.  In particular, they focus on retrieving perspectives that are relevant to a claim and consider a number of tasks: identifying the stance of these perspectives with respect to the original claim; identifying equivalent perspectives; and, given a claim and a perspective, identifying evidence for this stance from the associated corpus.  They also show that the RoBERTa Large model can perform reasonably well on these tasks. This approach has a number of similarities with ours: their identification of equivalent perspectives is analogous to our definition of a viewpoint as a collection of congruent claims. They also identify perspectives on the basis of a given claim, which is similar to our approach that identifies the claims relevant to a given topic. Having said so, we would argue that the conceptual framework underlying our approach provides a more robust basis for the realization of news analytics solutions than the one adopted by Chen et al. In particular, in their approach, there is no explicit characterization of a topic and the process of identifying relevant perspectives is driven by the selection of a particular claim. Here, it could be argued that it is unclear what is the epistemological basis for distinguishing between a claim and a perspective.  More importantly, while a claim implies a topic, the lack of an explicit characterization of the current topic makes it difficult to map the space of relevant viewpoints.  Indeed, in contrast with our approach, there is no attempt in~\cite{chen2022toward} at providing such a characterization of the space of viewpoints. On the contrary, topics, claims and viewpoints play clearly distinctive roles in our framework, which is grounded on the formal analysis and ontology presented in~\cite{motta2025epistemology}. Another difference between our work and that reported by Chen et al. concerns the architecture of the classifier used for the task of identifying equivalent perspectives – i.e., in our framework, the task of classifying a claim with respect to the available set of viewpoints.  In our approach we combine LLM fine-tuning and a semantic representation of the actors, in contrast with Chen et al., who only consider model fine-tuning. 

The paper by Doan et al.~\cite{doan2024automatically} covers a number of text mining tasks, including “political viewpoint identification”.  In particular, they test the performance of encoder-decoder models specialised for the Norwegian language on this task. Although the results from their preliminary tests are encouraging, a key difference with our approach is that while we characterise viewpoints as congruent collections of claims, no such clustering is considered by them.  Hence, their task is more similar to what we would call claim extraction.  Another limitation of their paper is that the structure of claims is not clearly defined – e.g., it is not clear whether the association between utterance and actor is captured in their approach.   Hada et al.~\cite{hada2023} focus on measuring viewpoint diversity within discussions on the X social media platform and propose the \textit{fragmentation} metric~\cite{vrijenhoek2021} to assess users' exposure to alternative viewpoints. Their analysis of conversations on immigration reveals that nearly 70\% of users have minimal exposure to alternative viewpoints. However, unlike our research, which aims to model the set of all viewpoints relevant to a topic, their work focuses solely on quantifying viewpoint diversity. Trabelsi and Zaïane~\cite{trabelsi2018} present an unsupervised generative Topic-Viewpoint model that uses unigram frequencies to probabilistically associate viewpoints to a topic. A key aspect of this approach is that it takes advantage of interactions between different post creators on social media to identify contrastive opinions. While this is of course an interesting and clever heuristic in the context of social media, it also means that this approach is not directly applicable to our news scenario, where such explicit interactions are usually not available. In addition, while this approach offers flexibility by avoiding reliance on predefined topics, it appears limited to binary classifications, either in favour or against a particular position related to a topic (e.g., gun control, gay marriage). This restricts its ability to capture the nuanced range of viewpoints often present in debates. This limitation can also be found more broadly in approaches focusing on \textit{stance detection}~\cite{alturayeif2023systematic}.  In particular, in contrast with stance detection, the relation between an agent and a topic is in our approach parametrised with respect to a viewpoint.  For instance, while a stance detection approach may detect positions in favour or against immigration, our approach would aim to identify the range of possible viewpoints related to this topic %(see section \ref{sec:ViewpointIdentification}) 
and then classify each claim with respect to this set. In particular, as discussed later, our case study focusing on UK immigration shows that an immigration-related viewpoint does not necessarily imply a pro/against stance on immigration – e.g., see viewpoint “Immigration as a Management Issue” in Section~\ref{sec:ViewpointIdentification}.

Finally, Calvo Vilares and He~\cite{calvo2017detecting} adopt an unsupervised LDA-based approach that attempts to jointly identify topics and viewpoints. They also generate readable summaries of the main viewpoints, by identifying sentences associated with the most discriminative words in the relevant topic-viewpoint model. However, they use a rather syntactic (i.e., keyword-based) approach to modelling, which appears to be rather noisy. As a result, the quality of the generated summaries tends to vary significantly, often highlighting sentences that do not necessarily express a viewpoint, in the sense of providing a contrastive opinion.  In addition, in contrast with both our work and the approach by Doan et al., no attempt is made here to cluster congruent claims.

\subsection{Integrating LLMs and KGs for Text Classification}
%LLMs have demonstrated impressive capabilities in processing complex text, establishing themselves as the new state-of-the-art for several NLP tasks, including text classification~\cite{ibrahim2024,junior2024largelanguagemodelsnew}. 

LLMs have shown impressive capabilities in processing and analyzing textual data at scale, allowing for the accurate extraction and synthesis of fine-grained information from a wide range of sources, including research papers~\cite{bolanos2024}, patents~\cite{kosonocky2024mining}, medical records~\cite{omiye2024large}, scientific references~\cite{buscaldi2024citation}, social media posts~\cite{yang2024mentallama}, news articles~\cite{motta2024}, technical documentation~\cite{cascini2004natural}, legal texts~\cite{savelka2023unreasonable}, knowledge graphs~\cite{meloni2025exploring}, tourism-related materials~\cite{cadeddu2024optimizing}, and financial documents~\cite{li2023large,birti2025optimizing}. 
However, they are often criticised for their tendency to hallucinate~\cite{tonmoy2024} and fabricate facts~\cite{augenstein2023factuality,wang2024}, and they also struggle to learn long-tail knowledge~\cite{zhou2023deviltailslongtailedcode}.
To overcome these limitations, researchers are exploring solutions, such as, retrieval-augmented generation (RAG)~\cite{lewis2021retrievalaugmentedgenerationknowledgeintensivenlp}, which involve providing LLMs with reliable and verifiable information. To this purpose, numerous approaches have been proposed in recent years, which integrate LLMs with KGs~\cite{peng2023knowledge,jiang2024}. These methods typically facilitate the incorporation of KGs during the fine-tuning or inference phases~\cite{pan2024}. In the former case, KGs are embedded into the model's training process, to enhance its ability to acquire and internalize knowledge~\cite{liu2020k,cadeddu2024comparative}. In the latter case, KGs are usually utilised for knowledge retrieval, enabling improved access to domain-specific information~\cite{meloni2025exploring,wu2023retrieve}. The approach presented in this paper uses the KG both during the fine-tuning and inference phases.

%The proposed approach captures both what the paper refers to as `global' semantics from the LLM and the `local' semantics from the KG~\cite{jiang2024}. 

%agentic AI~\cite{han2024llmmultiagentsystemschallenges}, 
%A recent survey~\cite{ibrahim2024} explores the integration of LLMs with KGs to improve AI system interpretability and performance. The study categorises integration approaches into three main paradigms: KG-augmented LLMs, LLM-augmented KGs, and synergised frameworks. Each paradigm is analysed in terms of methodology, strengths, weaknesses, and practical applications including text classification. The findings emphasise how these integrations enhance real-time data analysis, decision-making, and innovation across multiple domains. Additionally, the paper discusses key evaluation metrics, benchmarks, and challenges such as scalability and computational costs while proposing potential solutions.

Yin et al.~\cite{10650933} propose a novel framework, IKLT (Input Embedding, Knowledge Base Retrieval, Large Language Model Text Augmentation, and Text Encoding), tailored for text classification tasks. This framework employs RAG to integrate LLMs with knowledge bases, thereby enhancing feature representation while maintaining interpretability. Their evaluation, conducted on four public datasets and one industrial dataset, highlights the effectiveness of the proposed approach. Similarly to our work, they utilize knowledge bases to enrich the interpretability of text by augmenting its semantic content. However, their focus is primarily on short text, whereas our approach is applied to more elaborated news articles.

%The approach by Soman et al.~\cite{soman2024} integrates a knowledge graph, using RAG approach to improve LLMs responses to biomedical queries. Specifically, they use a knowledge graph to extract relevant information about a disease and present it to the LLM in a natural language format. This forces the LLM to focus on the most pertinent data improving the accuracy of its responses. This is an interesting approach that shares similarities with our work, but it does not tackle the task of classification.

Pons et al.~\cite{pons2024} propose an entity disambiguation approach framed as a classification task. In this context, instead of simply relying on the LLM to define the entity type, they provide the LLM with a curated set of relevant classes from a KG. To an extent, this approach is similar to ours, as we incorporate author information extracted from Wikidata, to enhance the classification of claims. However, Pons et al. only employ the KG at inference time.

In the news domain, Giarelis and his team~\cite{giarelis2024unified} present a framework for automated fact-checking within the context of public discourse. Their framework leverages an LLM to analyse an input text and identify the entities included in it. Then, it extracts relevant statements from the KG, which are fed back into the LLM through a refined prompt. This work is relevant to our research, as it employs LLMs in a multi-stage pipeline and integrates a KG for determining the veracity of specific claims.

% With regard to customising LLMs through a KG-enhanced fine-tuning, has showed to be challenging due to the high cost attached to the operation. Indeed, fine-tuning big LLMs (e.g., ChatGPT or LLaMA) 

% In this way, the model is able to effectively capture both global semantics from the LLM and local semantics from the KG.

%In this paper, we address a classification task by combining LLMs and KGs within the news domain. Differently from previous research, our method classifies claims based on viewpoints, demonstrating how integrating KGs can enhance the effectiveness of viewpoint analysis.

%In sum, while considerable research combines LLMs and KGs on a variety of tasks and use cases, their application in the news domain, and specifically, to classifying claims according to viewpoints remains largely unexplored. This paper aims to address this gap by demonstrating how a KG-enhanced model can effectively facilitate the analysis of viewpoints.

\section{Approach}\label{sec:methodology}

In this section, we present our approach to identifying claims from a set of news articles, which pertain to a given topic, and mapping them to a comprehensive set of relevant viewpoints. 
The proposed approach facilitates the development of various analytics that can reveal how specific viewpoints evolve over time, which ideologies or political parties endorse them, and how these perspectives are represented across different news outlets. These analytics can be used to provide insights about fairness and bias in the media, e.g., by indicating to what extent a particular media outlet, or a media landscape as a whole, provides a balanced account of the debate around a particular topic. %As a result, this method provides a sophisticated snapshot of the evolving news discourse.

\begin{figure}[h!]
    \centering
    \includegraphics[width=0.90\textwidth]{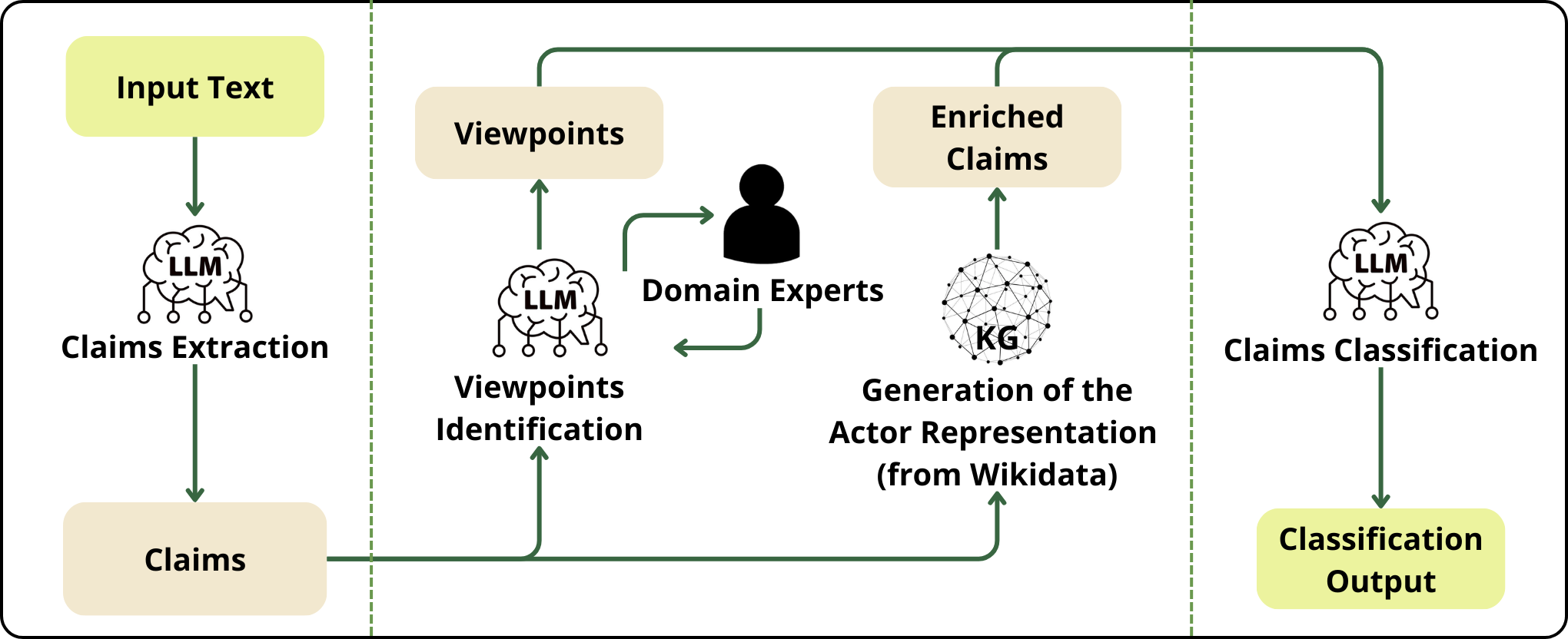}
    \caption{Methodology for claim extraction, enrichment, and classification.}
    \label{fig:esempio}
\end{figure}

%extends the pipeline originally introduced by Motta et al.~\cite{motta2024}, providing a significantly more sophisticated and effective approach to classifying claims. 

Our methodology, which extends the one originally presented in~\cite{motta2024}, is composed of four main steps —see Figure 1. First, a set of claims is extracted from news articles, where each claim consists of an utterance linked to a specific actor. Next, the viewpoints emerging from these claims are identified. In the third step, the representation of claims is enhanced by linking the relevant actors to an open KG. For this purpose, we adopted Wikidata, as it provides extensive coverage of public figures, politicians, and relevant groups, such as political parties. Finally, the claims are classified according to the viewpoints identified. As already mentioned, the key extensions to~\cite{motta2024} concern the following aspects: i) we enhance the representation of claims by acquiring information about the associated actors and ii) we fine-tune the LLM on the claim classification task.
%1) extracting a set of claims from news articles, consisting of an utterance associated with an actor; 2) identifying the primary viewpoints that emerge from these claims; 3) enhancing the representation of claims by linking the relevant actors to an open knowledge graph, such as Wikidata; and 4) classifying the claims based on the identified viewpoints. 

%The first two steps in the methodology have been explored in previous research~\cite{motta2024} and therefore only need to be briefly summarised here.  Hence, the discussion will focus primarily on the two downstream tasks in the pipeline: producing a semantic representation of the actors and classifying the claims.

%Figure 1 showcases the general architecture. It is structured into four main steps:

In the following subsections, we first discuss the background data required for this methodology and then outline the four steps.

\subsection{Background data}
The proposed framework processes a collection of news articles. Although it does not strictly depend on additional metadata, we highly recommend associating the articles with supplementary information to enhance subsequent analysis. The recommended metadata includes information about the relevant topics, location information, publication dates, and news outlets. This metadata is particularly valuable during two crucial stages. First, it aids in defining the scope of analysis, typically centred on topics (e.g., ``immigration'', ``Brexit'', ``human rights''), publication venues, geographic locations, or time periods. Second, it can be used to enrich the analysis once claims have been mapped to viewpoints. For instance, temporal data can illustrate how the prominence of certain viewpoints changes over time, while publication venues can reveal whether specific viewpoints are predominantly endorsed by particular newspapers.

%previously released by Motta et al.~\cite{motta2024}, which 

As in our earlier experiments ~\cite{motta2024}, we tested our pipeline on a dataset that includes news articles about immigration in the UK, published between June 1 and August 31, 2023. The corpus was constructed using the Aylien News API\footnote{Aylien - \url{https://aylien.com/}} and includes articles that were either pre-tagged with the keyword `immigration' by the Aylien service or contained this keyword, or closely related terms, such as `immigrants', in their title or body. %This dataset serves as a representative sample commonly employed by media studies experts to analyse how the media and various political actors frame and discuss the diverse perspectives on immigration. 
%In this paper, we use this dataset both as a running example and as the basis for the benchmark, which we use to evaluate the performance of our viewpoint classification methodology.

%We tested the methodology proposed in this paper on a dataset released by previous research~\cite{motta2024} consisting of 603 news articles about immigration in the UK, published between June 1 and August 31, 2023.

% Specifically, we retrieved news items that were either pre-tagged with the keyword immigration by the Aylien service or contained this keyword (or closely related terms, such as immigrants) in their title or body. The search was restricted to major mainstream news outlets, including The Guardian, BBC, ITV, Daily Mail, The Sun, The Daily Express, The Daily Telegraph, The Evening Standard, The Independent, Reuters, and Associated Press.

\subsection{Claim extraction and Viewpoint Identification}\label{sec:ViewpointIdentification}

%\color{purple}
%Add material from EKAW:

%To extract the set of claims we run GPT-4 on the entire news corpus, generating 3455 claims. However, once we restricted the corpus to UK immigration and the claims to utterances made by actors based in Britain (e.g., British politicians), we ended up with 766 entries. In addition, some of the utterances were essentially indistinguishable from one another and therefore we used Word Mover’s Distance [16] to identify duplicate utterances, utilizing a threshold value of 0.1. After this last step, the number of unique utterances was reduced to 589, even though the overall number of claims remained the same.
%\color{black}

The steps of claim extraction and viewpoint identification have already been carried out in our earlier experiments ~\cite{motta2024}, hence here we will only give a short overview of these tasks and we refer the reader to our earlier paper for more details. 

\subsubsection{Claim Extraction}

The objective of this step is to identify and extract a set of claims made by specific actors within a news article. A claim is defined as a statement attributed to an individual, as reported in the article. This can include both direct quotes, typically enclosed in quotation marks, as well as indirect statements, where the journalist paraphrases the original assertion (e.g., ``Labour's shadow home secretary Yvette Cooper has said the conservative government has `failed to get a grip' on channel crossings.''). %However, for simplicity, this paper focuses on directly reported claims, as they are easier to extract and verify.
In our implementation a claim consists of the following components: 1) a portion of text representing the utterance, 2) the name of the actor who made the statement, and 3) a link to the original article from which the claim was extracted. The metadata associated with the article then provide the other key components of a claim, i.e., news source, topic and date. Table 1 shows two claims extracted from our news corpus. % (the original articles have not been included for space purposes). 

% Table for claim components
\begin{table}[t]
    \centering
    % \scriptsize
    \caption{Examples of Claims. The links to the articles are in the footnotes. }
    \label{tab:claim_components}
    \begin{tabular}{>{\raggedright\arraybackslash}p{9cm}|>{\raggedright\arraybackslash}p{3cm}}
        \toprule
        \textbf{Utterance (URL news)} & \textbf{Actor} \\
        \midrule
        The Conservative government has `failed to get a grip' on Channel crossings. (\url{https://tinyurl.com/mr2wnbwf}) & Yvette Cooper \\
        \midrule
        Levels of migration are ‘too high’ and ‘unsustainable’. (\url{https://tinyurl.com/4ykb4tn2}) & Tom Hunt  \\
        \bottomrule
    \end{tabular}
\end{table}

In ~\cite{motta2024} we used GPT-4o with a carefully designed prompt\footnote{The prompt is available in the anonymous Github repository at \url{https://anonymous.4open.science/r/Political_Viewpoints_in_News_Media-3BE6/Prompts/Text_prompt_template.txt}} to extract claims and associated actors from the news corpus. The result was a set of 766 claims relevant to the UK immigration debate (plus many others associated with other topics), from which 402 were selected for our classification benchmark ~\cite{motta2024}.
These claims include concise, `atomic' statements, such as \textit{``Britain’s asylum system is riddled with abuse''} to more complex and nuanced assertions, such as, \textit{``Labour and the SNP repeatedly try to block our measures to control our borders, but where they're in power they refuse to accommodate their fair share of people who come here illegally. Time after time they play to the gallery and bang on about being `nations of sanctuary', but it's just warm words''}. % In reality, they're guilty of rank hypocrisy.'' %Rishi Sunak
%While measuring the performance of the claim extraction module is not a contribution of this paper, we note that the most advanced LLMs currently available, such as GPT-4, can execute the task with near-perfect accuracy. 

%This impressive performance can likely be attributed to our deliberate focus on directly reported claims enclosed in quotation marks, which LLMs can extract with relative ease. In future work, we intend to expand this approach to incorporate a broader range of indirect claims.

\subsubsection{Viewpoint Identification}

As discussed in ~\cite{motta2024}, this phase adopts a hybrid human–machine methodology to generate and refine viewpoints. The process leverages a large language model (GPT-4 Turbo in the current prototype) guided by carefully designed instructions to analyse a set of utterances and produce viewpoints relevant to the topic under discussion. 
The process unfolds in several stages. Initially, the utterances are partitioned into subsets small enough to be processed as input by the LLM. The model then analyses each subset, extracting a set of candidate viewpoints for that batch. Finally, a subsequent LLM prompt consolidates and refines the candidate viewpoints, ensuring coherence and relevance across the entire set. The resulting viewpoints are then reviewed and refined by human experts, whose primary role is to ensure that each viewpoint is clearly defined and distinct, and that the resulting collection of viewpoints is homogenous in terms of granularity and complete with respect to representing the perspectives relevant to the debate. To this purpose, where necessary, experts may introduce additional viewpoints that the automated process had not identified. This human-in-the-loop component is, in our view, essential. Our aim is to support researchers and practitioners engaged in media analytics for regulatory or research purposes and to enhance the human ability to analyse news media in a transparent manner, rather than fully automating the process. It is therefore crucial that the resulting set of viewpoints is both robust and consistent with the criteria that expert analysts, such as media and political scientists, would consider valid for analysing and classifying public debates.
In particular, the viewpoint identification step produced the following viewpoints relevant to the immigration news corpus ~\cite{motta2024}:

\begin{enumerate}
\item \textbf{Immigration as a management issue.} This viewpoint is used to classify utterances that take a `neutral' perspective on the immigration debate, focusing less on taking a pro- or anti-immigration stance than on discussing the advantages and disadvantages of specific approaches to managing the issue. This was the only viewpoint added manually by the domain expert. 
%This viewpoint characterises utterances that focus on the way immigration is managed, typically by the UK government. For example, criticisms of specific elements of immigration policy — e.g., the use of hotels to house immigrants — should be classified under this viewpoint, unless other factors, such as humanitarian considerations, are emphasised in the claim. A key aspect of this viewpoint is that it does not necessarily imply a stand in favour or against immigration.
\item \textbf{Immigrants as victims/Humanitarian emphasis.} %This viewpoint is used to classify utterances that are sympathetic to the plight of immigrants — e.g., when a tragedy happens at sea.
This viewpoint is used to classify utterances that are sympathetic to the plight of immigrants. For instance, when a tragedy happens at sea.
\item \textbf{Immigrants as potential criminals or threat/National security emphasis.} This viewpoint encompasses utterances that portray immigrants as a threat to law and order or even to national security.  %This includes statements referencing unethical lawyers assisting immigrants (implying criminal intent) and discussions of restrictive measures, such as the use of security tags, that suggest a perception of immigrants as inherently criminal.

%This viewpoint classifies utterances that imply a view of migrants as criminals or the migration phenomenon as a threat to national security. This viewpoint also covers the rhetoric about “dodgy lawyers” — i.e., lawyers who instruct their refugee clients to lie in order to get asylum in UK. In this case immigrants are criminals by association. In addition, utterances that advocate the use of restraining measures that are normally used for criminals — e.g., security tags, also fall under this category.
%(Any utterance that implies a view of immigrants as criminals; this includes statements that talk about dodgy lawyers (in this case immigrants are criminals because they get dodgy lawyers to assist them) and types of restrictions that imply a view of immigrants as criminals (e.g., use of security tags).)
\item \textbf{Enhancing/Maintaining immigration pathways.} This viewpoint includes both utterances advocating for measures that would facilitate easier entry into the UK and utterances that argue against imposing new restrictions to immigration.

%This viewpoint is used to classify utterances which either advocate for measures that would make it easier to come to UK or alternatively criticise the introduction of new restrictions to immigrations. Interestingly, in the context of an immigration debate, statements that criticise a relaxation of immigration rules should not be classified under ‘Maintaining immigration pathways’. In other words, ‘Maintaining immigration path-ways’ is not a neutral category, but implies a (mild) positive attitude towards immigration. In addition, this viewpoint is not necessarily mutually exclusive with the ‘Restricting immigration pathways’ one, because an utterance may advocate for more legal migrants coming to UK while supporting stricter measures against illegal migrants.
%(Any utterance which refers to measures that would make it easier to come to UK.)
\item \textbf{Restricting immigration pathways.} This viewpoint encompasses any utterance advocating for measures that would make it more difficult to enter the UK. %It addresses both legal and illegal immigration pathways and includes ``success stories'' about returning migrants to their home countries.

%This viewpoint is used to characterise utterances that refer to measures that would make it more difficult to come to UK. It covers both legal and illegal immigration pathways. Furthermore, attempts to remove migrants from the UK and ‘success stories’ about sending migrants back to their country also fall under this category.
%(Any utterance which refers to measures that would make it more difficult to come to UK; this dimension covers both legal and illegal immigration pathways; 'success stories' about sending migrants back to their country also fall under this category.)
\item \textbf{Economic benefits of immigration.} This viewpoint includes utterances highlighting the economic benefits that immigrants bring to the UK. 
%It excludes narrower remarks, such as those solely addressing increased visa costs.
%This viewpoint is used to classify utterances that refer to the economic value of immigration. Note that this viewpoint is not mutually exclusive with the one labelled ‘Economic cost of immigration’, be-cause an utterance may consider certain migrants as economically beneficial while maintaining that others introduce a financial burden for the country.
%(Utterances that refer to the economic value of immigration; these ought to be 'holistic' statements about the benefits of immigration, rather than, for example, statements about increasing visa costs.)
\item \textbf{Economic cost of immigration.} This viewpoint includes utterances that focus on the economic costs associated with immigration, such as expenses related to providing accommodation for undocumented migrants.

%This viewpoint is used to classify utterances that refer to the economic cost of immigration — e.g., when talking about the cost of accommodation for illegal migrants.
%(Utterances that refer to the economic cost of immigration - e.g., when talking about the cost of accommodation for illegal migrants.)
\item \textbf{Integration policies/Multiculturalism as a positive force.} This viewpoint includes utterances that suggest practical measures to support migrant integration or highlight the importance and value of cultural diversity.
%This viewpoint is used to classify utterances that propose practical measures for integrating migrants in UK society, emphasise the need to support the integration of migrants, or otherwise highlight the value of cultural diversity. This viewpoint is not mutually exclusive with the one labelled ‘Anti-integration policies/Cultural identity preservation’ because an utterance may express a favourable opinion about multiculturalism while advocating placing tracking tags on illegal migrants.
%(Utterances that propose practical measures to support the integration of migrants or emphasise the value of cultural diversity.)
\item \textbf{Anti-integration policies/Cultural identity preservation.} This viewpoint includes utterances that highlight cultural differences between UK citizens and migrants or advocate for the separation of migrants from the UK population.

%This viewpoint is used to classify utterances that emphasise the cultural differences between UK people and foreign migrants as well statements that advocate separating migrants from the rest of the UK population. For instance, the use of tracking tags on immigrants implies both a view of immigrants as criminals and also enforces an anti-integration policy.
%(Utterances that emphasise cultural differences between UK citizens and migrants or advocate for the separation of migrants from the UK population.)
\end{enumerate}

\color{black}

\subsection{Integrating the Semantic Representation of Actors}
%\subsection{Generating the Actor Knowledge Graph}

Most claims reported in the news are associated with significant figures in the media sphere, such as politicians, journalists, and other prominent individuals who frequently appear in the public eye. Information about these actors can often be obtained from various sources, including open knowledge graphs. In this work, we focused on Wikidata, as it typically provides extensive details about these individuals, including their political affiliations, occupations, roles in government, and past activities.

Enhancing claims with a semantic representation of the relevant actors derived from knowledge graphs serves two critical purposes. 
\textit{First}, it allows for the association of specific claims not only with individual persons, but also with broader categories, including political parties, governments, or ideologies. This capability is invaluable for identifying whether certain viewpoints are predominantly associated with specific groups, potentially uncovering patterns of bias in media coverage — e.g., over-representation of views from a specific political party in the news. % Moreover, this information can be utilised to generate insightful analytics, offering an intuitive understanding of which groups are likely to support specific viewpoints.
\textit{Second}, we hypothesise that the contextual information provided by a more detailed representation of the actors can improve the performance of machine learning classifiers in correctly associating claims to viewpoints. This is because statements are rarely made in isolation. They are often shaped by the speaker's history and identity, and interpreting them within this context provides valuable insights. In Section 6, we present evaluation results that strongly support this hypothesis, particularly when employing the largest models. 
Nevertheless, it should be noted that incorporating actor information may also introduce noise, especially when individuals make statements that deviate significantly from their typical positions. Striking a balance between leveraging an actor's history and avoiding the introduction of biases is a nuanced challenge, and we plan to explore it further in future work. 

To map actors to their corresponding Wikidata representations, we performed a targeted search using each actor's name to locate their associated entity on Wikidata. Once identified, we retrieved key information about each actor. In the current prototype, we consider their full name (\texttt{rdfs:label}), description (\texttt{schema:description}), gender (\texttt{P21: sex or gender}), occupations (\texttt{P106: occupation}), positions held (\texttt{P39: position held}), and political affiliations, such as membership of specific parties or groups (\texttt{P102: member of political party}). These attributes were identified empirically by analysing the most frequently populated and informative fields related to the actors in our data. Some attributes also support multiple values. For example, the `positions held' field may contain an extensive list of previous roles.

\begin{table}[h!]
% \scriptsize
\centering
\caption{Semantic representation of `Keir Starmer' (\url{https://www.wikidata.org/wiki/Q16515053}).\label{tab:generic_actor_semantic_representation}}
\resizebox{\textwidth}{!}{%
\begin{tabular}{l|l}
\toprule
\textbf{Field}                & \textbf{Value}                                                                                                                                     \\ \midrule
Name                          & Keir Starmer                                                                                                                                     \\ 
Gender                        & Male                                                                                                                                             \\ 
Description                   & Prime Minister of the United Kingdom since 2024        \\ 
Occupations                   & Barrister \\ 
                              & Politician \\ 
                              & Jurist \\ 
Political Party               & Labour Party \\
Positions Held                & Prime Minister of the United Kingdom \\ 
                              & Member of the 59th Parliament of the United Kingdom \\ 
                              & Leader of the Opposition \\  
                              & Leader of the Labour Party \\  
\bottomrule
\end{tabular}
}
\end{table}

In the immigration analysis used as our test case, we mapped the 52 actors associated with the 402 claims under analysis to their corresponding entities in Wikidata. For example, Table~\ref{tab:generic_actor_semantic_representation} presents the semantic representation of one such entity, `Keir Starmer', as extracted from Wikidata.

%The resulting semantic representation enriched the dataset, providing a robust foundation for analysing the relationships between actors and the claims they are associated with.
\color{black}

\subsection{Claim Classification}
Classifying a claim according to the relevant viewpoints is a complex task that requires a model capable of understanding abstract discourse. We formalise this as a binary classification problem: given an input claim, a context, and a description of the viewpoint, the model returns 1 if the claim aligns with the viewpoint and 0 otherwise. The claim includes both the utterance and the name of the actor. The context may include various types of information that facilitate the correct interpretation of the utterance. Specifically, we experimented with three different context settings. 
The first involves the surrounding text, including the sentences immediately preceding and following the utterance within the news article. The second employs text automatically derived from the set of triples from Wikidata describing the actor. Specifically, the triples are processed to generate a textual representation for each actor based on a simple template, making it more suitable for ingestion by the LLM for claim classification. For instance, the resulting textual description for Keir Starmer is \textit{``Keir Starmer (Prime Minister of the United Kingdom since 2024). He has worked as a barrister, politician, jurist and has held the position of Prime Minister of the United Kingdom, member of the 59th Parliament of the United Kingdom, Leader of the Opposition affiliated with the Labour Party with a religious or philosophical view of atheism and a political ideology of social democracy.
''}. The third and most comprehensive configuration combines both the surrounding text and the text derived from the triples. This final approach is currently adopted in our prototype as it delivered the best performance in our experiments, as detailed in Section 4. However, the increased length of this combined context is more computationally demanding and may also pose challenges, particularly for smaller models. We conducted the classification task using an LLM fine-tuned on a manually annotated dataset. Specifically, we utilized the Immigration-3K dataset, as described in Section 4.1. The prompt includes a task description, the claim to be analysed, the description of the viewpoint, and the relevant context. We evaluated several alternative models and configurations, identifying GPT-4o mini with the third type of context as the best-performing approach. This solution has now been integrated into our prototype.

\section{Evaluation}\label{sec:eval}
In this section, we present an evaluation of our claim classification approach, described in Sections 3.4 and 3.5, which is the key contribution of this paper. In particular, our main objective is to assess the value derived from i) enhancing claims through a semantic representation of the actors and ii) fine-tuning the models on high-quality data.

% 1) to demonstrate that this complex task, which requires understanding the ideological perspective underlying an utterance, can be effectively automated using modern AI technology; 2) 

%as the extraction of claims is performed with high accuracy using current LLMs, and the semi-automatic generation of viewpoints was discussed in previous work~\cite{motta2024}.

\subsection{Dataset: Immigration-3K}

%In this paper, we focus on evaluating the task of classifying a claim according to a viewpoint. This is the primary innovative task within our framework, as the extraction of claims is performed with high accuracy using current LLMs, and the semi-automatic generation of viewpoints has been addressed in previous work~\cite{motta2024}.

To set up a benchmark for this task, we reused the dataset originally introduced in Motta et al.~\cite{motta2024}. As already mentioned, this dataset includes 402 claims extracted from a news corpus covering the UK immigration debate. Five human annotators classified each claim against the nine relevant viewpoints detailed in Section 3.3, thus producing a benchmark of 10,854 claim-viewpoint data points, each annotated as a positive or negative match. However, it turned out that certain claims were potentially ambiguous and difficult to classify even for human experts. As a result, we found that the Cohen's kappa score on the full set of annotations was not robust enough, producing a "moderate" (0.42) average agreement score over the nine viewpoints ~\cite{landis1977measurement}. Therefore, we removed a set of 310 claim-viewpoint pairs, where one annotator classified the viewpoint as aligned with the claim, while the other two did not. This simple rule was sufficient to produce a robust dataset of 9,924 data points, with a substantial average agreement score of 0.66 across the nine viewpoints. We then applied majority rule over the three annotators to obtain a final dataset of 3,308 instances, named Immigration-3K, which is also available on GitHub\footnote{Immigration-3k is available in the anonymous GitHub repository at \url{https://anonymous.4open.science/r/Political_Viewpoints_in_News_Media-3BE6/Immigration-3K/test.csv}}.

Each row in the dataset includes a claim, a viewpoint description, and a boolean value, which indicates whether the viewpoint matches the claim. Each claim entry is associated with a unique ID, the utterance, the URL of the source article, the full text of the article, the name of the actor, and the description of the actor derived from Wikidata. To facilitate consistent use in future research, the dataset was pre-split into training, validation, and test sets, adhering to a 70\%-10\%-20\% ratio.
%containing 2,303, 336, and 669 rows, respectively, 

%\color{purple}
%FROM EKAW
%Our analysis also showed that viewpoint 8 (Integration policies/Multiculturalism as a positive force) was particularly problematic, with its opposite, viewpoint 9 (Anti-integration policies/Cultural identity preservation) also exhibiting a low agreement score. Therefore we also produced agreement scores with only seven viewpoints, re-moving viewpoints 8 and 9. 
%\color{black}

\subsection{Experimental setup}

We evaluated our methodology on the Immigration-3K benchmark, testing various LLMs and configurations. In particular, we are interested in evaluating a diverse range of LLMs, varying in nature and size. The experiments described in the paper employed the following LLMs:

\begin{itemize} \item \textbf{Gpt-4o mini}: A compact, task-optimised variant of GPT-4, offering a context size of 128k tokens\footnote{\url{https://platform.openai.com/docs/models\#gpt-4o-mini}}. Despite its smaller size, GPT-4-o mini outperforms the original GPT-4 in chat preferences. % and achieved an 82\% score on the Massive Multitask Language Understanding (MMLU) benchmark.

\item \textbf{Mistral Nemo 12B}: A model collaboratively developed by NVIDIA and Mistral AI. It supports a 128k token context window and excels in reasoning, world knowledge and coding accuracy, within its size category\footnote{\url{https://huggingface.co/nvidia/Mistral-NeMo-12B-Base}}.

\item \textbf{Gemma2 27B}: An open-source model from Google with a context size of 8k tokens. It has been trained on 13 trillion tokens sourced from diverse domains, including web documents, to ensure exposure to a wide range of linguistic styles and topics\footnote{\url{https://huggingface.co/google/gemma-2-27b}}.

\item \textbf{Gemma2 9B}: A smaller version of Gemma2 27B, trained on 8 trillion tokens\footnote{\url{https://huggingface.co/google/gemma-2-9b}}. It shares the same architecture and training data characteristics as its larger counterpart, making it suitable for deployment in resource-constrained environments. It also uses a context size of 8k tokens.

\item \textbf{Llama 3.1 8B}: A model developed by Meta, featuring a 128k-token context window\footnote{\url{https://huggingface.co/meta-llama/Llama-3.1-8B}}. Its architecture leverages cutting-edge advancements in transformer-based design, allowing it to process extensive contextual information efficiently, while maintaining high performance.

\item \textbf{Llama 3.2 3B}: A smaller version of Llama 3, featuring the same 128k token context size\footnote{\url{https://huggingface.co/meta-llama/Llama-3.2-3B}}.

\item \textbf{Phi 3.5 mini}: A model developed by Microsoft, featuring 3.8 billion parameters\footnote{\url{https://huggingface.co/microsoft/Phi-3.5-mini-instruct}}. It supports an extended context length of up to 128k tokens. It employs proximal policy optimization and direct preference optimization to ensure precise instruction adherence. 
\end{itemize}
We utilised the instruct variants of these models, which are specifically optimised using extensive datasets designed for instruction-based tasks. This pre-training process involved exposing the models to a diverse array of prompts accompanied by explicit instructions and corresponding outputs, significantly improving their capacity to comprehend and respond effectively to a wide range of instructional queries. We evaluated the model using two configurations: zero-shot learning (ZSL) and fine-tuning (FT). In the zero-shot setting, we employed a straightforward prompt to instruct the LLM to perform the classification task. For the fine-tuning approach, we utilised the training and validation splits of the Immigration-3K dataset.
The fine-tuning of Gpt-4o mini was conducted using the API provided by OpenAI\footnote{\url{https://platform.openai.com/docs/guides/fine-tuning}}, while the remaining models were fine-tuned with Unsloth\footnote{\url{https://docs.unsloth.ai/}}, an open-source library designed to optimize the fine-tuning and training of LLMs. %Unsloth is an open-source platform designed to optimize and accelerate the fine-tuning and training of LLMs.
%It offers significant speed improvements, enabling fine-tuning up to 2–5 times faster while reducing memory usage by up to 70\%.

While it is expected that fine-tuned models will outperform their zero-shot counterparts, including zero-shot evaluations serves two key purposes. First, it allows us to quantify the specific improvements gained from fine-tuning on high-quality, task-specific data. Second, it provides a baseline for comparison with previous experiments~\cite{motta2024}, which relied on off-the-shelf zero-shot methods.

Finally, as outlined in Section 3.5, our approach can adopt three types of context for each claim: 1) the text surrounding an utterance (labelled \textit{Text} in Table 3 and Table 4), 2) the representation of the actors from Wikidata (labelled \textit{KG}), and 3) both of them (labelled \textit{Text+KG}). We evaluate the performance of all models using these three contextual configurations. 

In conclusion, this experimental setup resulted in testing seven models in both zero-shot and fine-tuning configurations across three types of context, leading to a total of 42 experiments. 
Since this is a binary classification task that, given a claim and a viewpoint, produces 1 if the claim aligns with the viewpoint and 0 otherwise, we evaluated the performance using precision, recall, and F1 score.

% \begin{table}[h!]
% \centering
% \begin{tabular}{|c|c|c|c|c|c|c|c|c|c|}
% \hline
% \multirow{2}{*}{\textbf{Model}} & \multicolumn{3}{|c|}{\textbf{Article}} & \multicolumn{3}{|c|}{\textbf{Actor}} & \multicolumn{3}{|c|}{\textbf{Article-ActorFocus}} \\
% \cline{2-10}
%  & F1 & Precision & Recall & F1 & Precision & Recall & F1 & Precision & Recall \\
% \hline
% Gpt 4 o mini & 0.9233 & 0.9311 & 0.9160 & 0.8960 & 0.8864 & 0.9064 & 0.9033 & 0.9148 & 0.8928 \\
% \hline
% Mistral Nemo 12b & 0.8120 & 0.7945 & 0.8340 & 0.8378 & 0.8677 & 0.8145 & 0.7973 & 0.8237 & 0.7768 \\
% \hline
% Gemma2 9b & 0.8164 & 0.8192 & 0.8136 & 0.8223 & 0.8771 & 0.7864 & 0.8277 & 0.8481 & 0.8106 \\
% \hline
% Gemma2 27b & 0.8722 & 0.8791 & 0.8657 & 0.8593 & 0.9136 & 0.8221 & 0.8670 & 0.8703 & 0.8638 \\
% \hline
% Llama 3.1 8b & 0.7364 & 0.7364 & 0.7364 & 0.8517 & 0.8617 & 0.8425 & 0.6771 & 0.7200 & 0.6540 \\
% \hline
% Llama 3.2 3b & 0.6364 & 0.6740 & 0.6183 & 0.5190 & 0.8026 & 0.5300 & 0.5352 & 0.5338 & 0.5425 \\
% \hline
% Phi 3.5 mini & 0.6122 & 0.7817 & 0.5881 & 0.7925 & 0.8146 & 0.7749 & 0.6902 & 0.8194 & 0.6500 \\
% \hline
% \end{tabular}
% \caption{Comparison of F1, Precision, and Recall for Article, Actor, and Article-ActorFocus across various models using the Strict Dataset.}
% \label{tab:model_comparison_strict}
% \end{table}

%\vspace{-0.4cm}
\subsection{Results}

Table~\ref{tab:zsl_model_comparison} and Table~\ref{tab:model_comparison_maxCoverage} report the performance of the zero-shot models and the fine-tuned models, respectively. As expected, the fine-tuned models significantly outperform the zero-shot models.

\begin{table}[h!]
\centering
\scriptsize
\caption{F1, Precision, and Recall percentages of the seven models across the three context settings using a ZSL approach. The highest F1 score for each model is shown in bold, while the best overall result is highlighted in green.\label{tab:zsl_model_comparison}}
\resizebox{\textwidth}{!}{%
\begin{tabular}{l|c|c|c|c|c|c|c|c|c}
& \multicolumn{3}{c|}{\textbf{Text}} & \multicolumn{3}{c|}{\textbf{KG}} & \multicolumn{3}{c}{\textbf{Text+KG}} \\
\toprule
\textbf{Model} & F1 & Pre. & Rec. & F1 & Pre. & Rec. & F1 & Pre. & Rec. \\
\midrule
GPT-4o mini  & 77.80 & 73.41 & 88.12 & 80.10 & 76.50 & 85.92 & \cellcolor{green!30}\textbf{80.32} & 75.98 & 88.62 \\
\hline
Mistral Nemo 12B & 73.79 & 70.18 & 82.30 & \textbf{77.27} & 73.21 & 85.84 & 77.10 & 73.61 & 83.22 \\
\hline
Gemma2 27B & 71.13 & 68.05 & 85.43 & \textbf{73.21} & 69.52 & 85.00 & 72.01 & 68.57 & 83.91 \\
\hline
Gemma2 9B & 68.38 & 66.17 & 83.41 & \textbf{69.27} & 66.53 & 81.97 & 67.66 & 65.72 & 82.99 \\
\hline
Llama 3.1 8B & 63.51 & 63.16 & 79.46 & \textbf{71.50} & 68.17 & 81.20 & 68.04 & 65.43 & 78.42 \\
\hline
Llama 3.2 3B & 57.97 & 59.19 & 71.02 & 55.29 & 55.32 & 55.26 & \textbf{63.01} & 62.04 & 64.38 \\
\hline
Phi 3.5 mini & 59.16 & 60.92 & 76.01 & \textbf{73.73} & 70.25 & 81.28 & 62.39 & 62.14 & 76.84 \\
\bottomrule
\end{tabular}
}
\end{table}

The fine-tuned version of GPT-4o mini achieves the highest performance of all models (91.74\% F1), when utilising both the surrounding text and the actor description from Wikidata. This is followed by GPT-4o mini using only the actor description, and then by GPT-4o mini using only the surrounding text. This level of performance clearly improves on the results reported in~\cite{motta2024}, which correspond to the experiment where we used GPT-4o mini in a ZSL setting, with surrounding text as context. This experiment yielded an F1 score of 77.80\%.

We also note that several open LLMs perform well, even if they do not match the performance of GPT-4o mini.  
Notably, the fine-tuned versions of Mistral Nemo 12b (85.21\%), Gemma2 9B (84.78\%) and Gemma2 27B (84.13\%) achieved strong results. This demonstrates that it is possible to tackle this task effectively using open models, without requiring expensive hardware.

In addition, these results suggest that integrating information from an open KG can enhance performance, at least when leveraging an LLM capable of processing substantial input sizes. However, this improvement is not consistent across all tested LLMs. 
While GPT-4o mini achieved the best overall results when using the combined context, the other six tested LLMs exhibited mixed performance. Specifically, three models performed best when provided only with the actor description, while the remaining three performed better with the surrounding text. Interestingly, combining these two inputs led to a slight decrease in performance for these models. This could indicate that smaller LLMs struggle with larger inputs or with the differing nature of the two input sources.  Indeed, while many of the tested LLMs feature relatively large context windows, they do not necessarily have the ability to fully leverage all the relevant information~\cite{liu2024lost}. We plan to investigate this issue further in future research.

%Indeed why many of the tested LLMs have a relatively large context window, they dont necessary have the ability to fully utilise all the relevant information. 

The best zero-shot solution was the version of GPT-4o mini that utilised both the surrounding text and actor descriptions from Wikidata (80.32\%). However, this approach was still outperformed by all fine-tuned models, except for Llama 3.2 3B and Phi 3.5 mini. Notably, in all ZSL experiments, the solutions incorporating information from Wikidata, whether used alone or in combination with surrounding text, consistently outperformed those that relied solely on the surrounding text.

\begin{table}[h!]
\centering
\scriptsize
\caption{F1, Precision, and Recall percentages of the seven fine-tuned models across the three context settings. The highest F1 score for each model is shown in bold, while the best overall result is highlighted in green.\label{tab:model_comparison_maxCoverage}}
\resizebox{\textwidth}{!}{%
\begin{tabular}{l|c|c|c|c|c|c|c|c|c}
& \multicolumn{3}{c|}{\textbf{Text}} & \multicolumn{3}{c|}{\textbf{KG}} & \multicolumn{3}{c}{\textbf{Text+KG}} \\
\toprule
\textbf{Model} & F1 & Pre. & Rec. & F1 & Pre. & Rec. & F1 & Pre. & Rec. \\
\midrule
GPT-4o mini  & 90.30 & 93.13 & 87.93 & 90.33 & 91.08 & 89.62 & \cellcolor{green!30} \textbf{91.74} & 91.26 & 92.23 \\
\hline
Mistral Nemo 12B & \textbf{85.21} & 86.10 & 84.38 & 83.84 & 86.22 & 81.85 & 81.46 & 84.50 & 79.06 \\
\hline
Gemma2 27B & \textbf{84.13} & 89.54 & 80.32 & 82.28 & 90.96 & 77.20 & 82.81 & 85.68 & 80.50 \\
\hline
Gemma2 9B & 84.55 & 88.55 & 81.51 & \textbf{84.78} & 91.12 & 80.49 & 82.79 & 89.65 & 78.38 \\
\hline
Llama 3.1 8B & 78.62 & 87.87 & 73.65 & \textbf{81.65} & 83.37 & 80.16 & 73.31 & 71.76 & 75.27 \\
\hline
Llama 3.2 3B & \textbf{62.97} & 69.25 & 60.48 & 48.38 & 69.53 & 50.59 & 53.54 & 63.54 & 53.21 \\
\hline
Phi 3.5 mini & 67.83 & 70.72 & 65.89 & \textbf{76.03} & 86.66 & 70.95 & 57.91 & 85.97 & 55.91 \\
\bottomrule
\end{tabular}
}
\end{table}

Table~\ref{tab:byviewpoint} reports the performance of the best version of each model from Table~\ref{tab:model_comparison_maxCoverage} across the nine viewpoints outlined in Section 3.3. For most viewpoints, the models' performance closely aligns with their average performance. However, certain viewpoints present specific challenges, particularly for the smaller model. For instance, viewpoint 2, ``Humanitarian emphasis'', is handled reasonably well by GPT-4o mini but shows a marked decline in performance for the smaller models relative to their average results.

A few viewpoints are significantly more difficult to classify for all models. 
A manual analysis 
suggests that this difficulty arises from the dataset's inherent imbalance, with some viewpoints receiving extensive coverage, while others appear only sporadically. 
%A manual analysis indicates that this difficulty stems primarily from the fact that the dataset is highly unbalanced: while certain viewpoints are covered intensively in the news, others only appear sporadically.
This phenomenon, though natural in this domain and potentially indicative of imbalances in news coverage or in the discourse around a topic, can also impact the performance of the models in the claim classification task. 
%While from a domain point of view, this phenomenon appears to indicate a certain lack of balance and fairness in news coverage, it also affects the performance of the claim classification engine. 
This is particularly evident for viewpoints 6 (``Economic benefits of immigration'') and 8 (``Multiculturalism as a positive force''), which are heavily under-represented in our benchmark. 

%The political climate in the UK at the time the news articles were written contributed to this issue, as even left-leaning politicians were reluctant to make explicitly positive statements about these topics. Instead, they tended more often to criticise the Conservative Party's management of immigration, aligning more closely with viewpoint 1, ``Immigration as a management issue''.
%Consequently, these rare viewpoints were underrepresented in the dataset and, when present, were articulated in rather muted terms, limiting the ability of automated methods to classify them accurately. This finding underscores an interesting dynamic that merits further investigation in future work.
%\vspace{-0.2cm}
\begin{table}[h!]
\centering
% \scriptsize
\caption{F1 scores for the best-performing experiments in relation to the 9 viewpoints. In bold the best results for each viewpoint. \label{tab:byviewpoint}}
\resizebox{\textwidth}{!}{%
\begin{tabular}{l|c|c|c|c|c|c|c|c|c}
\toprule
\textbf{Model} & \textbf{1} & \textbf{2} & \textbf{3} & \textbf{4} & \textbf{5} & \textbf{6} & \textbf{7} & \textbf{8} & \textbf{9} \\
\midrule
GPT-4o mini (Text+KG) & 92.17 & \textbf{90.97} & \textbf{92.28} & 89.66 & \textbf{92.41} & 49.38 & 49.01 & \textbf{49.68} & \textbf{90.89} \\
\hline
Mistral Nemo 12B (Text) & \textbf{92.67} & 79.02 & 84.90 & \textbf{100} & 78.71 & \textbf{49.69} & 49.34 & \textbf{49.68} & 58.44 \\
\hline
Gemma2 27B (Text) & 90.52 & 70.52 & 76.74 & \textbf{100} & 81.80 & \textbf{49.69} & 49.67 & \textbf{49.68} & 47.76 \\
\hline
Gemma2 9B (KG) & 87.79 & 82.31 & 74.70 & 89.66 & 84.62 & \textbf{49.69} & \textbf{83.00} & \textbf{49.68} & 62.41 \\
\hline
Llama 3.2 3B (Text) & 54.22 & 62.61 & 62.05 & 49.33 & 54.81 & \textbf{49.69} & 47.62 & \textbf{49.68} & 47.37 \\
\hline
Llama 3.1 8B (KG) & 82.90 & 70.52 & 82.66 & 49.33 & 84.83 & 49.38 & 68.99 & 49.35 & 55.60 \\
\hline
Phi 3.5 mini (KG) & 67.55 & 62.61 & 54.66 & 83.00 & 78.61 & \textbf{49.69} & 49.67 & \textbf{49.68} & 62.41 \\
\bottomrule
\end{tabular}
}
\end{table}
%\vspace{-1cm}

\section{Conclusions}\label{sec:conclusions}

%In this paper, we propose a novel hybrid human–machine methodology that integrates LLMs with KGs to identify and analyse diverse viewpoints within news corpora. This approach advances the state of the art in the automatic analysis of news through AI and represents a step toward the development of agentic applications capable of achieving a deeper understanding of news and its influence on public opinion.

In this paper, we have presented an approach to the task of classifying claims in news articles with respect to a given set of viewpoints, which improves over previous work on the automatic analysis of perspectives in the news. 
Our approach leverages LLMs fine-tuned on a high-quality dataset and employs a semantic characterisation of the relevant actors, namely the individuals who articulate a specific claim within a news piece. Each actor is mapped to a set of triples extracted from Wikidata, which provide information such as occupations, ideologies, historical roles, and other relevant attributes. This strategy extends the analytical capabilities of our method by enabling the association of viewpoints with specific groups or ideologies, while also providing rich contextual information for interpreting claims, thereby improving the overall performance of the approach. We evaluated our approach on Immigration-3K, a dataset that describes the mapping between 402 claims and 9 viewpoints concerning the immigration debate in the UK. The full version of our pipeline achieves excellent results and outperforms alternative solutions, with an F1 score of 91.74\%. The evaluation also suggests that introducing a semantic description of the actors enhances performance, particularly when leveraging an LLM capable of processing substantial input sizes. For our future research, we foresee several directions. First, we intend to conduct extensive experiments across multiple topics to evaluate how effectively a system trained on certain topics (e.g., immigration) can be reused for others (e.g., climate change, health policy). To this end, we also plan to release a new benchmark covering multiple topics, which will provide a highly valuable resource for researchers interested in news analytics. Furthermore, we aim to investigate how to enhance the system’s ability to address rarer viewpoints, for which it is harder to identify a sufficient number of examples. In particular, we will experiment with the generation of synthetic data and assess whether this solution can help address this issue. Finally, we intend to develop a complete application that can be used by both political scientists and other user audiences to produce analytics across multiple topics and countries. Indeed, our ultimate goal is to develop a robust and practical tool capable of assisting a variety of users in analysing media representations of democratic discourse.

%\appendix
%\section{Example Appendix Section}
%\label{app1}

%%       \cite{<label>} ==> [1]

%%\medskip

%\textbf{Supporting Information} \par %Please delete the Suppporting Information statement if it is not applicable. Please supply Supporting Information in another file. Supporting information should not be provided in .tex format
%Supporting Information is available from the Wiley Online Library or from the author.

% Acknowledgements
%\medskip
%\textbf{Acknowledgements} \par %delete if not applicable))
%Please insert your acknowledgements here

% References
\medskip

% Use the following code if you wish to generate your bibliography with BibTeX;
% replace the string "MSP-template" below with the name(s) of
% the BibTeX data base(s) you want to use.
% The resulting bibliography-output (the content of the .bbl file)
% must be pasted back into this file before submission.
% Please also include your BibTeX data base file(s) in your submission
% so that we can re-run BibTeX if necessary.
%
\bibliographystyle{MSP}
\bibliography{bibliography}

@article{cadeddu2024comparative,
  title={A comparative analysis of knowledge injection strategies for large language models in the scholarly domain},
  author={Cadeddu, Andrea and Chessa, Alessandro and De Leo, Vincenzo and Fenu, Gianni and Motta, Enrico and Osborne, Francesco and Recupero, Diego Reforgiato and Salatino, Angelo and Secchi, Luca},
  journal={Engineering Applications of Artificial Intelligence},
  volume={133},
  pages={108166},
  year={2024},
  publisher={Elsevier}
}

@article{bolanos2024,
	author = {Bola{\~n}os, Francisco and Salatino, Angelo and Osborne, Francesco and Motta, Enrico},
	da = {2024/08/17},
	doi = {10.1007/s10462-024-10902-3},
	id = {Bola{\~n}os2024},
	isbn = {1573-7462},
	journal = {Artificial Intelligence Review},
	number = {10},
	pages = {259},
	title = {Artificial intelligence for literature reviews: opportunities and challenges},
	ty = {JOUR},
	url = {https://doi.org/10.1007/s10462-024-10902-3},
	volume = {57},
	year = {2024}}

@inproceedings{liu2020k,
  title={K-bert: Enabling language representation with knowledge graph},
  author={Liu, Weijie and Zhou, Peng and Zhao, Zhe and Wang, Zhiruo and Ju, Qi and Deng, Haotang and Wang, Ping},
  booktitle={Proceedings of the AAAI conference on artificial intelligence},
  volume={34},
  pages={2901--2908},
  year={2020}
}

@article{wu2023retrieve,
  title={Retrieve-rewrite-answer: A kg-to-text enhanced llms framework for knowledge graph question answering},
  author={Wu, Yike and Hu, Nan and Bi, Sheng and Qi, Guilin and Ren, Jie and Xie, Anhuan and Song, Wei},
  journal={arXiv preprint arXiv:2309.11206},
  year={2023}
}

@inproceedings{li2023large,
  title={Large language models in finance: A survey},
  author={Li, Yinheng and Wang, Shaofei and Ding, Han and Chen, Hang},
  booktitle={Proceedings of the fourth ACM international conference on AI in finance},
  pages={374--382},
  year={2023}
}

@article{cadeddu2024optimizing,
  title={Optimizing tourism accommodation offers by integrating language models and knowledge graph technologies},
  author={Cadeddu, Andrea and Chessa, Alessandro and De Leo, Vincenzo and Fenu, Gianni and Motta, Enrico and Osborne, Francesco and Reforgiato Recupero, Diego and Salatino, Angelo and Secchi, Luca},
  journal={Information},
  volume={15},
  number={7},
  pages={398},
  year={2024},
  publisher={MDPI}
}

@article{peng2023knowledge,
  title={Knowledge graphs: opportunities and challenges},
  author={Peng, Ciyuan and Xia, Feng and Naseriparsa, Mehdi and Osborne, Francesco},
  journal={Artificial Intelligence Review},
  pages={1--32},
  year={2023},
  publisher={Springer}
}

@article{meloni2025exploring,
  title={Exploring Large Language Models for Scientific Question Answering via Natural Language to SPARQL Translation},
  author={Meloni, Antonello and Reforgiato Recupero, Diego and Osborne, Francesco and Salatino, angelo and Motta, Enrico and Vahadati, Sahar and Lehmann, Jens},
  journal={ACM Transactions on Intelligent Systems and Technology},
  year={2025},
  publisher={ACM New York, NY}
}

@inproceedings{cascini2004natural,
  title={Natural language processing of patents and technical documentation},
  author={Cascini, Gaetano and Fantechi, Alessandro and Spinicci, Emilio},
  booktitle={International Workshop on Document Analysis Systems},
  pages={508--520},
  year={2004},
  organization={Springer}
}

@article{savelka2023unreasonable,
  title={The unreasonable effectiveness of large language models in zero-shot semantic annotation of legal texts},
  author={Savelka, Jaromir and Ashley, Kevin D},
  journal={Frontiers in Artificial Intelligence},
  volume={6},
  pages={1279794},
  year={2023},
  publisher={Frontiers Media SA}
}

@article{birti2025optimizing,
  title={Optimizing Large Language Models for ESG Activity Detection in Financial Texts},
  author={Birti, Mattia and Osborne, Francesco and Maurino, Andrea},
  journal={arXiv preprint arXiv:2502.21112},
  year={2025}
}

@article{kosonocky2024mining,
  title={Mining patents with large language models elucidates the chemical function landscape},
  author={Kosonocky, Clayton W and Wilke, Claus O and Marcotte, Edward M and Ellington, Andrew D},
  journal={Digital Discovery},
  volume={3},
  number={6},
  pages={1150--1159},
  year={2024},
  publisher={Royal Society of Chemistry}
}

@article{omiye2024large,
  title={Large language models in medicine: the potentials and pitfalls: a narrative review},
  author={Omiye, Jesutofunmi A and Gui, Haiwen and Rezaei, Shawheen J and Zou, James and Daneshjou, Roxana},
  journal={Annals of internal medicine},
  volume={177},
  number={2},
  pages={210--220},
  year={2024},
  publisher={American College of Physicians}
}

@inproceedings{yang2024mentallama,
  title={MentaLLaMA: interpretable mental health analysis on social media with large language models},
  author={Yang, Kailai and Zhang, Tianlin and Kuang, Ziyan and Xie, Qianqian and Huang, Jimin and Ananiadou, Sophia},
  booktitle={Proceedings of the ACM Web Conference 2024},
  pages={4489--4500},
  year={2024}
}

@article{buscaldi2024citation,
  title={Citation prediction by leveraging transformers and natural language processing heuristics},
  author={Buscaldi, Davide and Dess{\'\i}, Danilo and Motta, Enrico and Murgia, Marco and Osborne, Francesco and Recupero, Diego Reforgiato},
  journal={Information Processing \& Management},
  volume={61},
  number={1},
  pages={103583},
  year={2024},
  publisher={Elsevier}
}

@article{liu2024lost,
  title={Lost in the middle: How language models use long contexts},
  author={Liu, Nelson F and Lin, Kevin and Hewitt, John and Paranjape, Ashwin and Bevilacqua, Michele and Petroni, Fabio and Liang, Percy},
  journal={Transactions of the Association for Computational Linguistics},
  volume={12},
  pages={157--173},
  year={2024},
  publisher={MIT Press One Broadway, 12th Floor, Cambridge, Massachusetts 02142, USA~…}
}

@article{augenstein2023factuality,
  title={Factuality challenges in the era of large language models},
  author={Augenstein, Isabelle and Baldwin, Timothy and Cha, Meeyoung and Chakraborty, Tanmoy and Ciampaglia, Giovanni Luca and Corney, David and DiResta, Renee and Ferrara, Emilio and Hale, Scott and Halevy, Alon and others},
  journal={arXiv preprint arXiv:2310.05189},
  year={2023}
}

@inproceedings{calvo2017detecting,
  title={Detecting perspectives in political debates},
  author={Calvo Vilares, David and He, Yulan},
  booktitle={Proceedings of the Conference on Empirical Methods in Natural Language Processing},
  pages={1574--1583},
  year={2017},
  organization={Association for Computational Linguistics}
}

@INPROCEEDINGS{10650933,
  author={Yin, Haoran},
  booktitle={2024 International Joint Conference on Neural Networks (IJCNN)}, 
  title={An Industrial Short Text Classification Method Based on Large Language Model and Knowledge Base}, 
  year={2024},
  volume={},
  number={},
  pages={1-7},
  doi={10.1109/IJCNN60899.2024.10650933}}

@misc{zhou2023deviltailslongtailedcode,
      title={The Devil is in the Tails: How Long-Tailed Code Distributions Impact Large Language Models}, 
      author={Xin Zhou and Kisub Kim and Bowen Xu and Jiakun Liu and DongGyun Han and David Lo},
      year={2023},
      eprint={2309.03567},
      archivePrefix={arXiv},
      primaryClass={cs.SE},
      url={https://arxiv.org/abs/2309.03567}, 
}

@article{alturayeif2023systematic,
  title={A systematic review of machine learning techniques for stance detection and its applications},
  author={Alturayeif, Nora and Luqman, Hamzah and Ahmed, Moataz},
  journal={Neural Computing and Applications},
  volume={35},
  number={7},
  pages={5113--5144},
  year={2023},
  publisher={Springer}
}

@inproceedings{doan2024automatically,
  title={Automatically Detecting Political Viewpoints in Norwegian Text},
  author={Doan, Tu My and Baumgartner, David and Kille, Benjamin and Gulla, Jon Atle},
  booktitle={International Symposium on Intelligent Data Analysis},
  pages={242--253},
  year={2024},
  organization={Springer}
}

@article{trabelsi2018, title={Unsupervised Model for Topic Viewpoint Discovery in Online Debates Leveraging Author Interactions}, volume={12}, url={https://ojs.aaai.org/index.php/ICWSM/article/view/15021}, DOI={10.1609/icwsm.v12i1.15021}, abstractNote={ &lt;p&gt; Online debate forums provide a valuable resource for textual discussions about controversial social and political issues. Discovering the viewpoints and their discourse or arguments from such resources is important for policy and decision makers. In order to detect the stance, most of the existing methods rely on expensively obtained human annotations and propose supervised solutions. In this work, we introduce a purely unsupervised Author Interaction Topic Viewpoint model (AITV) for viewpoint identification at the post and the discourse levels. The model favors &quot;heterophily&quot; over &quot;homophily&quot; when encoding the nature of the authors’ interactions in online debates. It assumes that the difference in viewpoints breeds interactions, unlike similar studies based on social network analysis, which hypothesize that similar viewpoints encourage interactions. We evaluate the model’s viewpoint identification and clustering accuracies at the author and post levels. Experiments are held on six corpora about four different controversial issues, extracted from two online debate forums. AITV’s results show a better performance in terms of viewpoint identification at the post level than the state-of-the-art supervised methods in terms of stance prediction, even though it is unsupervised. It also outperforms a recently proposed topic model for viewpoint discovery in social networks and achieves close results to a weakly guided unsupervised method in terms of author level viewpoint identification. Our results highlight the importance of encoding &quot;heterophily&quot; for purely unsupervised viewpoint identification in the context of online debates. We also carry out a brief qualitative evaluation of the discourse modeling in terms of Topic-Viewpoint word clusters. AITV shows encouraging results suggesting an accurate discovery of the viewpoints and topics’ discourses. &lt;/p&gt; }, number={1}, journal={Proceedings of the International AAAI Conference on Web and Social Media}, author={Trabelsi, Amine and Zaiane, Osmar}, year={2018}, month={Jun.} }

@inproceedings{hada2023,
author = {Hada, Rishav and Ebrahimi Fard, Amir and Shugars, Sarah and Bianchi, Federico and Rossini, Patricia and Hovy, Dirk and Tromble, Rebekah and Tintarev, Nava},
title = {Beyond Digital "Echo Chambers": The Role of Viewpoint Diversity in Political Discussion},
year = {2023},
isbn = {9781450394079},
publisher = {Association for Computing Machinery},
address = {New York, NY, USA},
url = {https://doi.org/10.1145/3539597.3570487},
doi = {10.1145/3539597.3570487},
booktitle = {Proceedings of the Sixteenth ACM International Conference on Web Search and Data Mining},
pages = {33–41},
numpages = {9},
location = {Singapore, Singapore},
series = {WSDM '23}
}

@misc{tonmoy2024,
      title={A Comprehensive Survey of Hallucination Mitigation Techniques in Large Language Models}, 
      author={S. M Towhidul Islam Tonmoy and S M Mehedi Zaman and Vinija Jain and Anku Rani and Vipula Rawte and Aman Chadha and Amitava Das},
      year={2024},
      eprint={2401.01313},
      archivePrefix={arXiv},
      primaryClass={cs.CL},
      url={https://arxiv.org/abs/2401.01313}, 
}

@misc{jiang2024,
      title={KG-FIT: Knowledge Graph Fine-Tuning Upon Open-World Knowledge}, 
      author={Pengcheng Jiang and Lang Cao and Cao Xiao and Parminder Bhatia and Jimeng Sun and Jiawei Han},
      year={2024},
      eprint={2405.16412},
      archivePrefix={arXiv},
      primaryClass={cs.CL},
      url={https://arxiv.org/abs/2405.16412}, 
}

@InProceedings{pons2024,
author="Pons, Gerard
and Bilalli, Besim
and Queralt, Anna",
editor="Demartini, Gianluca
and Hose, Katja
and Acosta, Maribel
and Palmonari, Matteo
and Cheng, Gong
and Skaf-Molli, Hala
and Ferranti, Nicolas
and Hern{\'a}ndez, Daniel
and Hogan, Aidan",
title="Knowledge Graphs for Enhancing Large Language Models in Entity Disambiguation",
booktitle="The Semantic Web -- ISWC 2024",
year="2025",
publisher="Springer Nature Switzerland",
address="Cham",
pages="162--179",
isbn="978-3-031-77844-5"
}

@misc{wang2024,
      title={Factuality of Large Language Models: A Survey}, 
      author={Yuxia Wang and Minghan Wang and Muhammad Arslan Manzoor and Fei Liu and Georgi Georgiev and Rocktim Jyoti Das and Preslav Nakov},
      year={2024},
      eprint={2402.02420},
      archivePrefix={arXiv},
      primaryClass={cs.CL},
      url={https://arxiv.org/abs/2402.02420}, 
}

@inproceedings{vrandevcic2023wikidata,
  title={Wikidata: The making of},
  author={Vrande{\v{c}}i{\'c}, Denny and Pintscher, Lydia and Kr{\"o}tzsch, Markus},
  booktitle={Companion Proceedings of the ACM Web Conference 2023},
  pages={615--624},
  year={2023}
}

@InProceedings{motta2024,
author="Motta, Enrico
and Osborne, Francesco
and Pulici, Martino M. L.
and Salatino, Angelo
and Naja, Iman",
editor="Alam, Mehwish
and Rospocher, Marco
and van Erp, Marieke
and Hollink, Laura
and Gesese, Genet Asefa",
title="Capturing{\thinspace}the Viewpoint Dynamics in the News Domain",
booktitle="Knowledge Engineering and Knowledge Management",
year="2025",
publisher="Springer Nature Switzerland",
address="Cham",
pages="18--34",
isbn="978-3-031-77792-9"
}

@article{motta2025epistemology,
  title={The epistemology of fine-grained news classification},
  author={Motta, Enrico and Daga, Enrico and Gangemi, Aldo and Gjelsvik, Maia Lunde and Osborne, Francesco and Salatino, Angelo},
  journal={Semantic Web},
  volume={16},
  number={3},
  pages={22104968251344461},
  year={2025},
  publisher={SAGE Publications Sage UK: London, England}
}

@ARTICLE{pan2024,
  author={Pan, Shirui and Luo, Linhao and Wang, Yufei and Chen, Chen and Wang, Jiapu and Wu, Xindong},
  journal={IEEE Transactions on Knowledge and Data Engineering}, 
  title={Unifying Large Language Models and Knowledge Graphs: A Roadmap}, 
  year={2024},
  volume={36},
  number={7},
  pages={3580-3599},
  doi={10.1109/TKDE.2024.3352100}}

@misc{lewis2021retrievalaugmentedgenerationknowledgeintensivenlp,
      title={Retrieval-Augmented Generation for Knowledge-Intensive NLP Tasks}, 
      author={Patrick Lewis and Ethan Perez and Aleksandra Piktus and Fabio Petroni and Vladimir Karpukhin and Naman Goyal and Heinrich Küttler and Mike Lewis and Wen-tau Yih and Tim Rocktäschel and Sebastian Riedel and Douwe Kiela},
      year={2021},
      eprint={2005.11401},
      archivePrefix={arXiv},
      primaryClass={cs.CL},
      url={https://arxiv.org/abs/2005.11401}, 
}

@article{landis1977measurement,
  title={The Measurement of Observer Agreement for Categorical Data},
  author={Landis, JR},
  journal={Biometrics},
  year={1977}
}

@article{Masini18112018,
author={Masini, Andrea and Van Aelst, Peter and Zerback, Thomas and Reinemann, Carsten and Mancini, Paolo and Mazzoni, Marco and Damiani, Marco and Coen, Sharon},
title = {Measuring and Explaining the Diversity of Voices and Viewpoints in the News},
journal = {Journalism Studies},
volume = {19},
number = {15},
pages = {2324--2343},
year = {2018},
publisher = {Routledge},
doi = {10.1080/1461670X.2017.1343650},
URL = { 
        https://doi.org/10.1080/1461670X.2017.1343650
},
eprint = {    
        https://doi.org/10.1080/1461670X.2017.1343650
}
}

@inproceedings{giarelis2024unified,
  title={A Unified LLM-KG Framework to Assist Fact-Checking in Public Deliberation},
  author={Giarelis, Nikolaos and Mastrokostas, Charalampos and Karacapilidis, Nikos},
  booktitle={Proceedings of the First Workshop on Language-driven Deliberation Technology (DELITE)@ LREC-COLING 2024},
  pages={13--19},
  year={2024}
}

@book{trappel2011media,
  title={The Media for Democracy Monitor: A cross national study of leading news media},
  author={Trappel, Josef and Nieminen, Hannu and Nord, Lars W},
  year={2011},
  publisher={Nordicom, University of Gothenburg}
}

@article{Smith01072012,
author = {Rachael Craufurd Smith and Damian Tambini},
title = {Measuring Media Plurality in the United Kingdom: Policy Choices and Regulatory Challenges},
journal = {Journal of Media Law},
volume = {4},
number = {1},
pages = {35--63},
year = {2012},
publisher = {Routledge},
doi = {10.5235/175776312802483862},
URL = { 
        https://doi.org/10.5235/175776312802483862
},
eprint = {     
        https://doi.org/10.5235/175776312802483862
}
}

@book{chen2022toward,
    title = "Creating a More Transparent Internet: The Perspective Web",
    editor = "Piek Vossen and Antske Fokkens",
    year = "2022",
    doi = "10.1017/9781108641104",
    language = "English",
    isbn = "9781108485760",
    publisher = "Cambridge University Press",
}

@inproceedings{vrijenhoek2021,
author = {Vrijenhoek, Sanne and Kaya, Mesut and Metoui, Nadia and M\"{o}ller, Judith and Odijk, Daan and Helberger, Natali},
title = {Recommenders with a Mission: Assessing Diversity in News Recommendations},
year = {2021},
isbn = {9781450380553},
publisher = {Association for Computing Machinery},
address = {New York, NY, USA},
url = {https://doi.org/10.1145/3406522.3446019},
doi = {10.1145/3406522.3446019},
booktitle = {Proceedings of the 2021 Conference on Human Information Interaction and Retrieval},
pages = {173–183},
numpages = {11},
location = {Canberra ACT, Australia},
series = {CHIIR '21}
}

\end{document}